\documentclass[sigconf]{acmart}
\AtBeginDocument{%
  }

\usepackage{color}
\usepackage{graphicx}
\usepackage{textcomp}
\usepackage{xcolor}
\usepackage{url}
\usepackage{multirow}
\usepackage{times}
\usepackage{latexsym}
\usepackage{todonotes}
\usepackage{multirow}
\usepackage{graphicx}
\usepackage{amsmath}
\usepackage{amsfonts}
\usepackage{caption}
\usepackage[T1]{fontenc}
\usepackage{pifont}
\usepackage{subcaption}
\usepackage{booktabs}
\usepackage{colortbl}
\usepackage{tikz}
\usepackage{xcolor}

\definecolor{goldcolor}{RGB}{184,134,11}
\definecolor{lightbeige}{RGB}{245,245,220}
\definecolor{lightgray}{RGB}{211,211,211}
\definecolor{lightmint}{RGB}{176,224,230}
\definecolor{darkmint}{RGB}{96,168,154}

\usepackage{enumitem}
\usepackage{pifont}
\usepackage{graphicx}

\usepackage[utf8]{inputenc}

\definecolor{LightCyan}{rgb}{0.88,1,1}
\usepackage{graphicx}
\usepackage{subcaption}
\usepackage{enumitem}
\usepackage{xcolor}
\usepackage{color, colortbl}
\usepackage[utf8]{inputenc}
\usepackage{times}
\usepackage{latexsym}
\usepackage{comment}
\usepackage[utf8]{inputenc}
\usepackage{subcaption}
\usepackage[T1]{fontenc}
\usepackage{microtype}
\usepackage{algpseudocode}
\usepackage{times}
\usepackage{latexsym}
\usepackage{pgfplots}
\usepackage{svg}
\usepackage{makecell}
\usepackage{tikz}
\usetikzlibrary{positioning}
\usetikzlibrary{backgrounds}
\usepackage{tikzscale}
\usepackage{pgfplots}
\pgfplotsset{width=10cm,compat=1.9}
\usepgfplotslibrary{groupplots}
\usetikzlibrary{pgfplots.statistics}
\usetikzlibrary{patterns}
\usepackage{subcaption}
\usepackage{multirow}
\usepackage{adjustbox}
\usepackage{enumitem}
\usepackage{comment}
\usepgfplotslibrary{groupplots}
\usepackage[T1]{fontenc}
\definecolor{g1}{rgb}{0,0.8,0.4}
\definecolor{g4}{rgb}{0.88,1,0.88}
\definecolor{g2}{rgb}{0.66,1,0.66}
\definecolor{g3}{rgb}{0.8,1,0.8}

\definecolor{r1}{rgb}{1.0, 0.03, 0.0}
\colorlet{r2}{r1!50}
\colorlet{r3}{r1!30}
\colorlet{r4}{r1!15}

\begin{document}

\title{ChartCitor: Answer Citations for ChartQA via Multi-Agent LLM Retrieval}

\author{Kanika Goswami}
\affiliation{%
  \institution{IGDTUW, Delhi}
  \country{India}}
\author{Puneet Mathur}
\affiliation{%
  \institution{Adobe Research}
  \country{USA}}
\author{Ryan Rossi}
\affiliation{%
  \institution{Adobe Research}
  \country{USA}}
\author{ Franck Dernoncourt}
\affiliation{%
  \institution{Adobe Research}
  \country{USA}}
  
\begin{abstract}
Large Language Models (LLMs) can perform chart question-answering tasks but often generate unverified hallucinated responses. Existing answer attribution methods struggle to ground responses in source charts due to limited visual-semantic context, complex visual-text alignment requirements, and difficulties in bounding box prediction across complex layouts. We present \texttt{ChartCitor}, a multi-agent framework that provides fine-grained bounding box citations by identifying supporting evidence within chart images. The system orchestrates LLM agents to perform chart-to-table extraction, answer reformulation, table augmentation, evidence retrieval through pre-filtering and re-ranking, and table-to-chart mapping. \texttt{ChartCitor} outperforms existing baselines across different chart types. Qualitative user studies show that \texttt{ChartCitor} helps increase user trust in Generative AI by providing enhanced explainability for LLM-assisted chart QA and enables professionals to be more productive.
\end{abstract}




\maketitle

\section{Introduction}

Chart data finds extensive use across diverse domains such as healthcare, finance, and education. Recently, LLMs such as Llama-3.2 \cite{touvron2023llama}, Claude-3.5 Sonnet, and GPT-4V \cite{openai} have proven effective in utilizing in-context learning and visual prompting to interpret and reason over chart images. However, these models tend to hallucinate — generate answers with semantically plausible but factually incorrect information — which undermines their reliability and erodes user trust \cite{Xu2024HallucinationII, Snyder2023OnED}. While existing approaches attempt to address hallucination by grounding LLM-generated responses in source documents through citation mechanisms \cite{ji2023survey}, charts present unique challenges: (i) complex mapping between visual elements and underlying data, (ii) limited contextual information due to compressed visual data representation, (iii) difficulty in localizing chart elements across diverse visualization types and layouts, and (iv) ambiguity in alignment between text descriptions and visual elements. Prior research has explored various approaches to address this challenge, including instruction tuning \cite{hagrid}, in-context learning \cite{gao-etal-2023-enabling}, and natural-language inference (NLI)-based post-hoc attribution methods \cite{gao-etal-2023-rarr}. However, these approaches have primarily focused on attributing entire charts rather than specific structural elements \cite{huo2023retrieving}, limiting their practical utility. To address these limitations, we propose \texttt{ChartCitor}, a system that provides visual evidence for generated answers by identifying and highlighting relevant chart elements through bounding box annotations. \texttt{ChartCitor} works by orchestrating multiple specialized LLM agents to: (1) extract structured data table from charts, (2) break down answers into logical steps, (3) generate contextual descriptions for rows/columns, (4) identify supporting evidence through pre-filtering and re-ranking to connect specific table cells to claims, and (5) localize the selected cells in the chart image. \texttt{ChartCitor} helps professionals save time on fact-checking LLM-generated answers and enhances user trust by providing reliable and logically-explained citations sourced from charts.

\begin{figure*}
\centering
\scalebox{1}{
\includegraphics[width=0.9\textwidth]{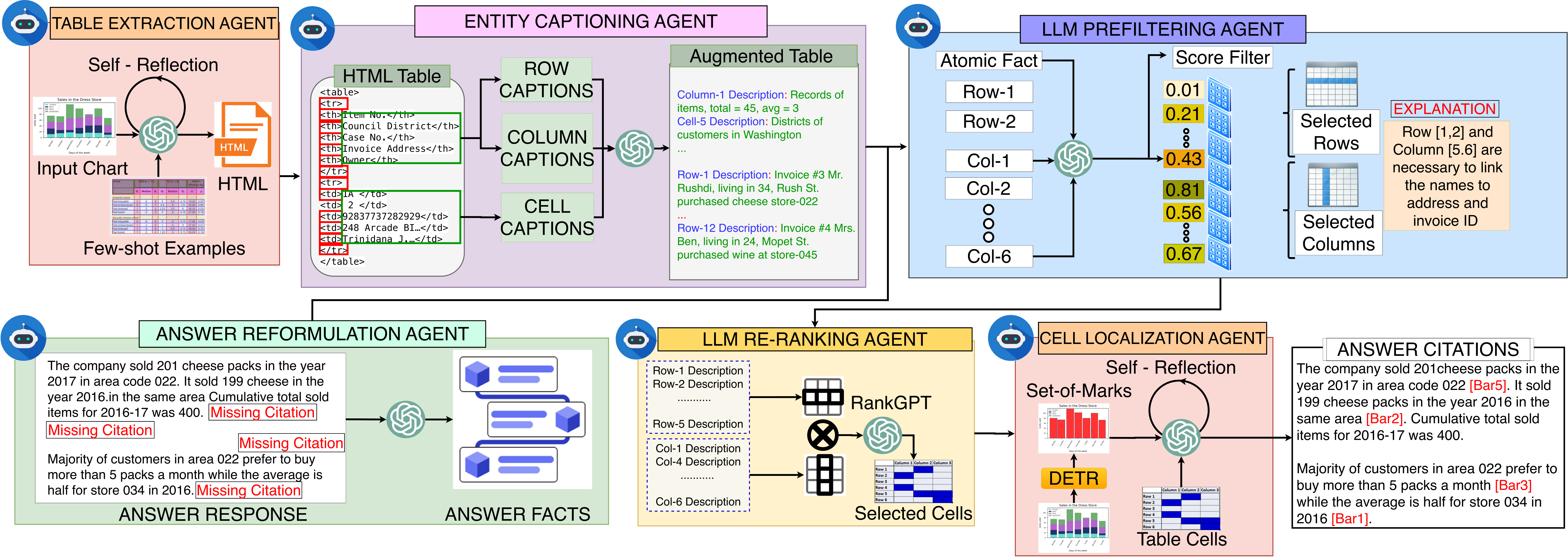}}
\caption{\small\texttt{ChartCitor} - a multi-agent framework that performs table extraction, answer reformulation, entity captioning, row/col retrieval, and cell localization in chart images to ground answers.}
\label{fig:main}
\end{figure*}

\section{ChartCitor}

We aim to solve the Fine-grained Structured Chart Attribution task which involves identifying graph elements (e.g bars, lines, pies in chart images) that support factual claims in a generated text response to a user's question. We propose \texttt{ChartCitor} (Fig. \ref{fig:main}) -- a multi-agent framework that provides fine-grained citations for generated answers grounded in chart image by orchestrating multiple LLM agents, which is explained as follows:

\noindent\textbf{(1) Chart2Table Extraction Agent}: Charts are predominantly present in PDFs or scanned documents that need to be converted into structured table formats (e.g., CSV, or HTML). We utilize GPT-4V to comprehend PDF images and output corresponding HTML without the need for external OCR using few shot prompting to identify cell data across each row/column. We use visual self-reflection \cite{Shinn2023ReflexionAA} to provide the GPT-4V with its own rendered HTML and data table output to check for consistency between the re-plotted LLM output and the original chart. In case of inconsistencies in the data extraction, the LLM refines its output until the extracted table data is error-free.

\noindent\textbf{(2) Answer Reformulation Agent}: The answer to be attributed, which can be AI-generated or otherwise, may be composed of multiple facts, numerical formulations and multi-hop logic. Each fact may be sourced from a different row/column in the chart table. To facilitate precise citations, re-framing the answer statement into a chain of reasoning steps helps to better retrieve the correct citations from the table. We convert the answer statement into a hierarchy of reasoning thoughts/arguments via few-shot in-context prompting, ensuring the resultant answer arguments are independent sentences without any  deviation in their collective meaning from the original answer statement.

\noindent\textbf{(3) Entity Captioning Agent}: Understanding tabular data extends beyond simple cell interpretation, requiring comprehension of how cell information relate to both the table's structure and its broader context. Tables often present analytical challenges through ambiguous content, including technical terminology, contextless numeric values, domain-specific symbols, and hierarchical row/column  headers. These ambiguities impede reliable evidence extraction and citation validation through semantic matching. Our solution leverages LLMs in an unsupervised manner to generate rich, multi-layered contextual descriptions: \textbf{(i) Row Captioning}: Our system employs GPT-4o to generate comprehensive row-level descriptions that capture complex patterns across features, summarize temporal trends, highlight significant dates and provide comparative analysis with respect to outliers within each row. \textbf{(ii) Column Captioning}: We generate detailed captions for each column using GPT-4o to explain ambiguous measurement units, symbols, empty cell spaces, and technical relationship of it's contents with corresponding row headers. \textbf{(iii) Cell Captioning}: Row and column-level captions may highlight broader trends but the fine-grained cell level information needs to be contextualized in terms of its associated row and column headers. Captioning agent uses GPT-4o to describe the importance of each cell in the context of its associated row and columns.

\noindent\textbf{Table Cell Retrieval}: We use retrieve-then-rank approach to identify the most relevant table cells. Our two-step approach begins with LLM-based pre-filtering to reject irrelevant rows and columns. We then employ LLM re-ranking to retrieve the most precise cell-level matches, ensuring both comprehensive coverage and accuracy in the final selection.

\noindent\textbf{(4) LLM Pre-filtering Agent}: We hypothesize that some of the table rows/ columns are likely to be unrelated to the answer facts. Passing irrelevant and distracting table entities to the LLM-based re-ranker can mislead it, negatively impacting the ranking process. Inspired by \cite{nouriinanloo2024re}, the LLM-based pre-filtering step uses chain-of-thought \cite{Wei2022ChainOT} followed by Plan and Solve \cite{Wang2023PlanandSolvePI} prompting techniques to generate a relevance score for each row/column based on the significance of its descriptive caption to the given answer statement (between $0$ to $1$). Additionally, we prompt the LLM to explain its rationale behind the score generation to enhance explainability and avoid hallucinations. We establish a specific threshold (usually $0.3 - 0.5$) for row/column filtering to retain potential citations that are sent to the re-ranker, discarding those falling below the threshold. This implementation significantly reduces the number of noisy of rows and columns that can misguide the re-ranker, leading to an improved citation retrieval performance.

\begin{figure*}[ht]
    \centering
    
    \begin{minipage}[c]{0.48\textwidth}
        \centering
        \scriptsize
        \begin{tabular}{l|c}
            \toprule
            \bf Method & \bf IoU  \\ \hline
            Kosmos-2 & 3.89 \\
            LISA & 4.34 \\
            GPT-4V (Direct Bbox Decoding) & 12.5 \\ 
            Claude-3.5 (Sonnet Direct Bbox Decoding) & 13.8 \\ 
            DETR \cite{carion2020end}+ Set-of-Marks Prompting \cite{Yang2023SetofMarkPU} & 18.6 \\ \hline
            \rowcolor{g3} \texttt{ChartCitor} & \bf 27.4 \\
            \hline
        \end{tabular}
        \label{tab:ablation_results}
    \end{minipage}%
    \begin{minipage}[c]{0.48\textwidth}
        \centering
        \includegraphics[width=0.8\textwidth]{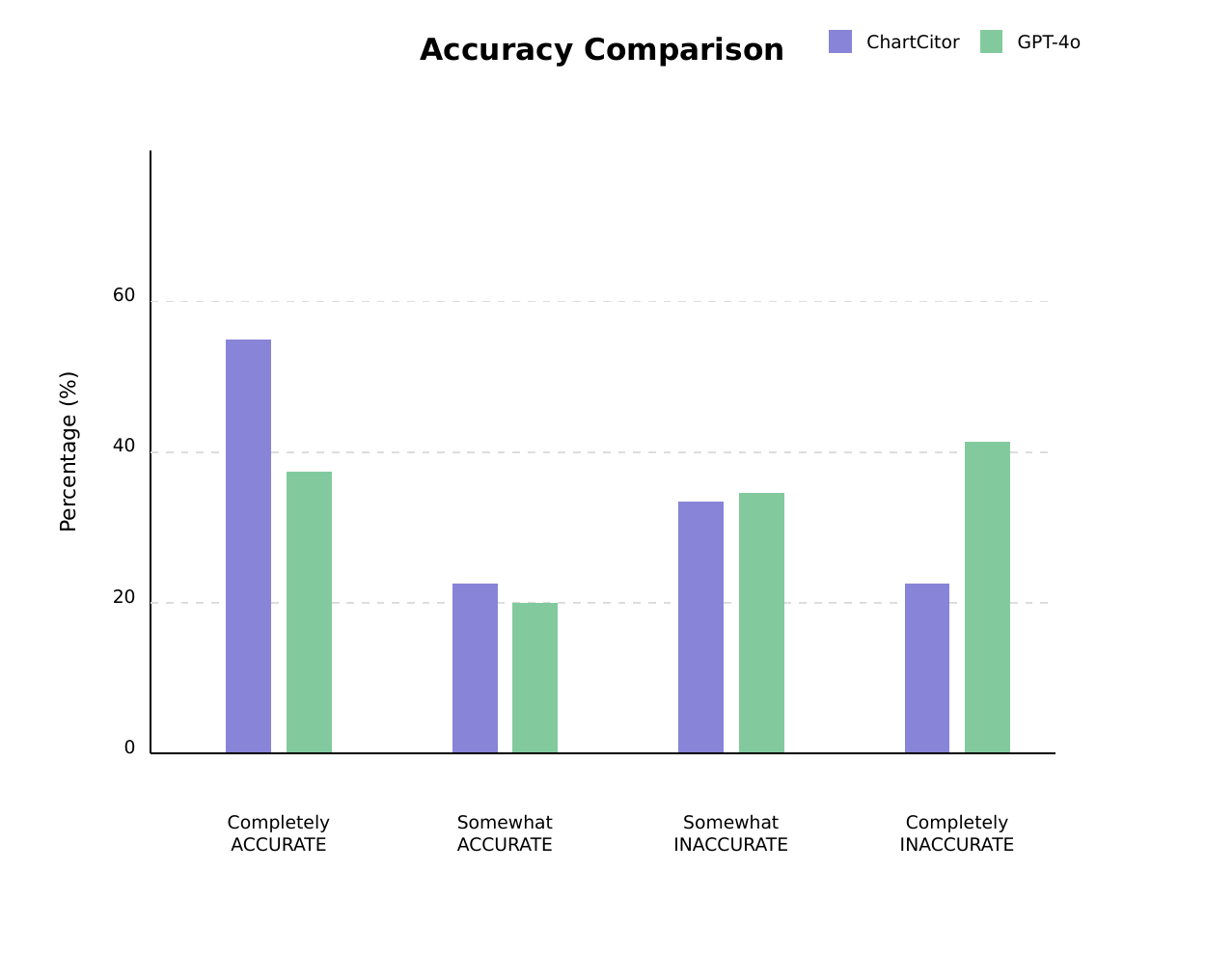}
        \label{fig:plot2}
    \end{minipage}
    \caption{\label{tab:ablation_results} (a) Ablation analysis of multimodal feedback agents; (b) User Evaluation of ChartCitor}
\end{figure*}

\noindent\textbf{(5) LLM Re-ranking Agent}: LLMs providing citations that don't directly support their claims can be interpreted as a form of hallucination which may diminish user confidence. To solve this, we retrieve the set of table cells that are collectively both sufficient and directly relevant to the answer claims. We use RankGPT \cite{sun-etal-2023-chatgpt}, a listwise LLM re-ranker to re-rank the table cells extracted from the intersection of rows and columns selected in the pre-filtering stage. Additionally, we prompt GPT-4o to provide a layer-of-thought \cite{fungwacharakorn2024layer} explanation of its rationale in ranking the cell items, enhancing the transparency in the citation mechanism by enforcing a logical coherence in the evidence chain.

\noindent\textbf{(6) Cell Localization Agent}: The final step maps cited table cells to their corresponding visual elements in the chart image. This agent leverages DETR trained \cite{carion2020end} on ChartQA data to identify all possible data marks (bars, line segments, pie slices) using image processing algorithms. GPT-4V with few-shot set-of-marks prompting \cite{yang2023set} then identifies elements corresponding to the cited cells. The agent generates bounding box coordinates for the relevant visual elements, employing visual self-reflection to verify precise correspondence between highlighted regions and cited data points. 

\section{Implementation Details}

\noindent All constituent agents utilize textual LLM APIs such as those provided by OpenAI (GPT-4o, GPT-4V) or Claude Sonnet-3.5. We use \texttt{ChatGPT-4o} as the base multimodal language model (MLLM) for ChartCitor. We convert data tables from TabCite benchmark \cite{Mathur2024MATSAMT} into bar/pie/line charts along with paired QA.

\noindent\textbf{Evaluation}: We adopt visual Intersection over Union (IoU) as principal metric for chart attribution tasks. Detected regions in the chart image are matched to ground truth regions (e.g., bars in barplot or pies in piechart) based on a threshold value of $ \text{IoU} \geq 0.9 $. Unlike bar charts and pie charts, where detected regions can be matched to discrete ground truth regions, line charts involve discrete points. Since grounding models generate bounding boxes or regions, we compute the proportion of ground truth points covered within the detected region(s) over total points detected.

\noindent\textbf{Baselines}: (1) Zero-shot LLM Bounding Box Prompting -- We prompt GPT-4o and Claude 3.5 Sonnet to predict normalized bounding box coordinates for chart components based on input text and the visual chart. (2) Kosmos-2 \cite{peng2023kosmos} is a multimodal LLM with text-to-visual grounding capabilities. It represents object locations as Markdown links for generating bounding boxes for visual grounding tasks. (3) LISA (Large Language Instructed Segmentation Assistant) \cite{lai2024lisa} is a reasoning-based zero-shot segmentation model that generates masks from implicit and complex textual queries.

\section{Results and Discussion}

Table \ref{tab:ablation_results}(a) shows quantitative results that demonstrate that \texttt{ChartCitor} consistently outperforms the baselines across all chart types, highlighting its robustness and effectiveness in visual chart understanding. \texttt{ChartCitor} achieves better performance compared to directly prompting LLMs to predict bounding boxes. Further, even using GPT-4V with set of marks prompting over detected chart elements show weak performance. Kosmos-2 and LISA perform poorly, with very low IoU scores, highlighting their inability to handle factual grounding in charts due to insufficient visual and numerical reasoning. Interestingly, all tested method including our proposed \texttt{ChartCitor}, zero-shot LMM prompting, LISA and KOSMOS2 struggle with interpreting complex geometrical proportions in pie charts due to their difficulty in handling non-rectangular bounding box segmentation task. Further, we conducted a user study (Fig. \ref{tab:ablation_results}(b) to evaluate the citation accuracy and perceived utility of fine-grained chart attribution provided by \texttt{ChartCitor}. Five participants evaluated 250 randomly sampled question-answer pairs with associated chart images to study the usefulness and accuracy of the citations provided by \texttt{ChartCitor} compared to direct GPT-4o prompting. The evaluation results demonstrated strong positive reception, with participants rating the attributions as \textit{Completely Accurate} ($41\%$ vs $28\%$) or \textit{Somewhat Accurate} ($17\%$ vs $15\%$) for verifying chart-based question answering accuracy in ChartCitor and GPT-4o, respectively. Attributions were found to be more "Completely Inaccurate" ChartCitor than GPT-4o ($17\%$ vs $31\%$). Participants described the citations as a handy tool in making verification of LLM-generated answers easier (\textit{"...can help me to quickly verify trends in charts, cutting down the time I spent on 10 documents from 5 hrs to 20 mins."}).


\section{Conclusion}

We introduced ChartCitor, which grounds LLM-generated QA responses to chart elements using agentic orchestration and set-of-marks prompting. The system outperforms baselines by 9-15\% and shows promise for rich document QA over PDF collections. While currently effective for single-chart citations, future work can address multi-chart interactions, hallucination mitigation, and explicit citation-text mapping to enhance trustworthy multimodal content generation.

\bibliographystyle{ACM-Reference-Format}
\bibliography{sample-base}


\end{document}